%% file: conference_041818.tex
\documentclass[conference,a4paper]{IEEEtran}
\IEEEoverridecommandlockouts
\usepackage{cite}
\usepackage{amsmath,amssymb,amsfonts}
\usepackage{algorithmic}
\usepackage{graphicx}
\usepackage{textcomp}
\usepackage{multirow}
\usepackage{caption}
\usepackage{subcaption}
\usepackage{xcolor}
\usepackage[final]{pdfpages}
\def\BibTeX{{\rm B\kern-.05em{\sc i\kern-.025em b}\kern-.08em
    T\kern-.1667em\lower.7ex\hbox{E}\kern-.125emX}}

\begin{document}

\title{Multiple Document Datasets Pre-training Improves Text Line Detection With Deep Neural Networks}

\author{\IEEEauthorblockN{M\'elodie Boillet}
\IEEEauthorblockA{\textit{LITIS, Université de Rouen-Normandie} \\
\textit{TEKLIA, Paris}\\
Rouen, France \\
boillet@teklia.com}
\and
\IEEEauthorblockN{Christopher Kermorvant}
\IEEEauthorblockA{\textit{LITIS, Université de Rouen-Normandie} \\
\textit{TEKLIA, Paris}\\
Rouen, France \\
kermorvant@teklia.com}
\and
\IEEEauthorblockN{Thierry Paquet}
\IEEEauthorblockA{\textit{LITIS, Université de Rouen-Normandie} \\
Rouen, France \\
}

}

\maketitle

\begin{abstract}
In this paper, we introduce a fully convolutional network for the document layout analysis task. While state-of-the-art methods are using models pre-trained on natural scene images, our method Doc-UFCN relies on a U-shaped model trained from scratch for detecting objects from historical documents. We consider the line segmentation task and more generally the layout analysis problem as a pixel-wise classification task then our model outputs a pixel-labeling of the input images. We show that Doc-UFCN outperforms state-of-the-art methods on various datasets and also demonstrate that the pre-trained parts on natural scene images are not required to reach good results. In addition, we show that pre-training on multiple document datasets can improve the performances. We evaluate the models using various metrics to have a fair and complete comparison between the methods.
\end{abstract}


\section{Introduction}
\input{parts/01-introduction}

\section{Related works}
\label{related_work}
\input{parts/02-related_works}

\section{Model}
\label{model} 
\input{parts/03-model}

\section{Data}
\label{data} 
\input{parts/04-data}

\section{Comparison to state-of-the-art}
\label{comparison_SOTA} 
\input{parts/05-comparison_sota}

\section{Ablation study}
\label{ablation_study} 
\input{parts/06-ablation_study}

\section{Conclusion}
\input{parts/07-conclusion}

\section*{Acknowledgments}
This work benefited from the support of the French National Research Agency (ANR) through the project HORAE ANR-17-CE38-0008. It is also part of the \textit{HOME History of Medieval Europe} research project supported by the European JPI Cultural Heritage and Global Change (Grant agreements No. ANR-17-JPCH-0006 for France, MSMT-2647\\/2018-2/2, id. 8F18002 for Czech Republic and PEICTI Ref. PCI2018-093122 for Spain). Mélodie Boillet is partly funded by the CIFRE ANRT grant No. 2020/0390. 

\bibliographystyle{IEEEtran}
\bibliography{references}


\end{document}

%% file: parts/01-introduction.tex
Automatic document understanding and more specifically the layout analysis of historical documents is still an active area of research. This task consists in splitting a document into different regions according to their content. It can be a very challenging task due to the variety of documents. In this paper, we focus on text line segmentation in historical documents.

Some recent works using pre-trained weights \cite{dhsegment} have emerged. The use of  pre-training has shown many advantages such as decreasing the training time and improving the model's accuracy. However, these weights are often learned on natural scene images (ImageNet \cite{ImageNet} dataset) and then applied to document images. Since document images are really different from natural scene images, in this paper we question the interest of using such a pre-training stage.

\begin{figure}[htbp]
    \centerline{
        \includegraphics[width=0.247\textwidth]{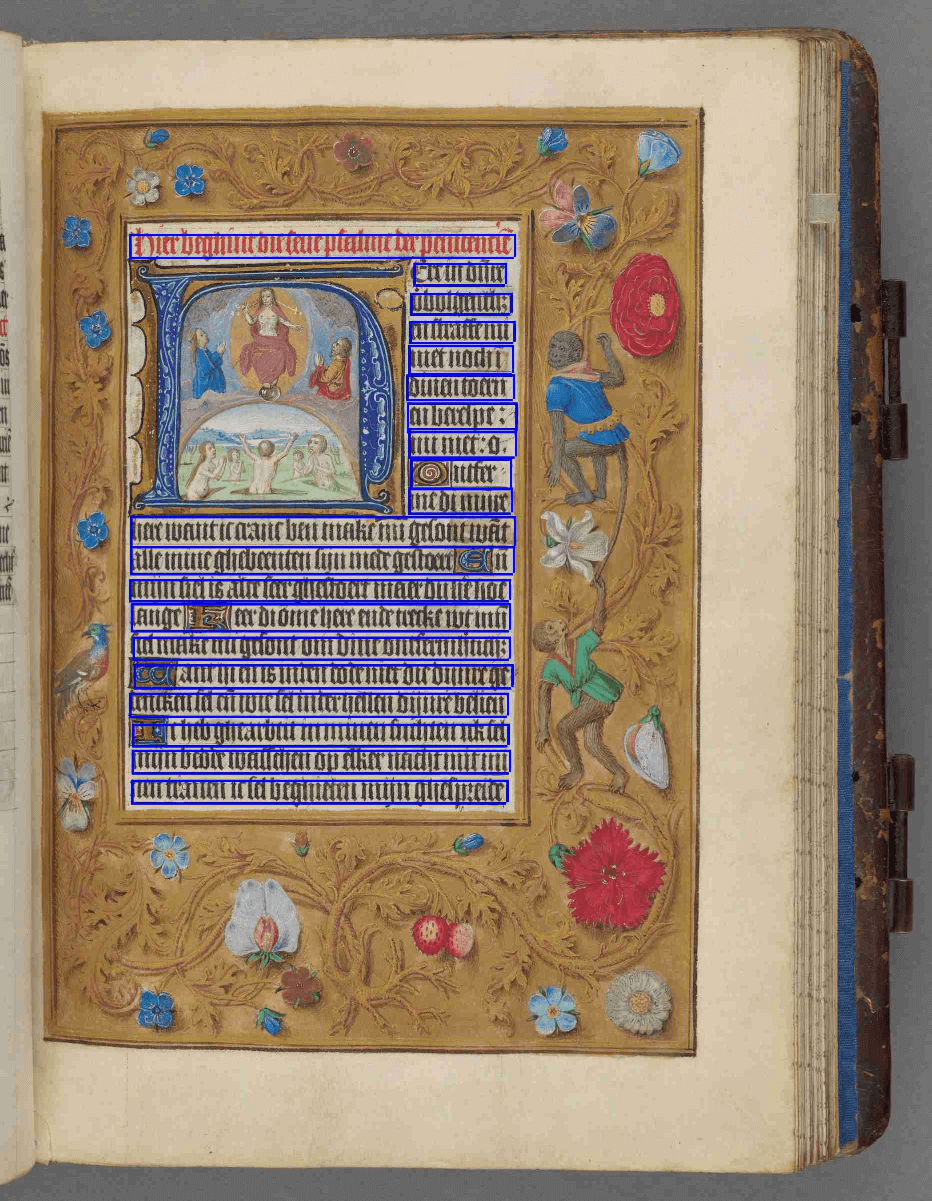}
        \hspace{0.1cm}
        \includegraphics[width=0.204\textwidth]{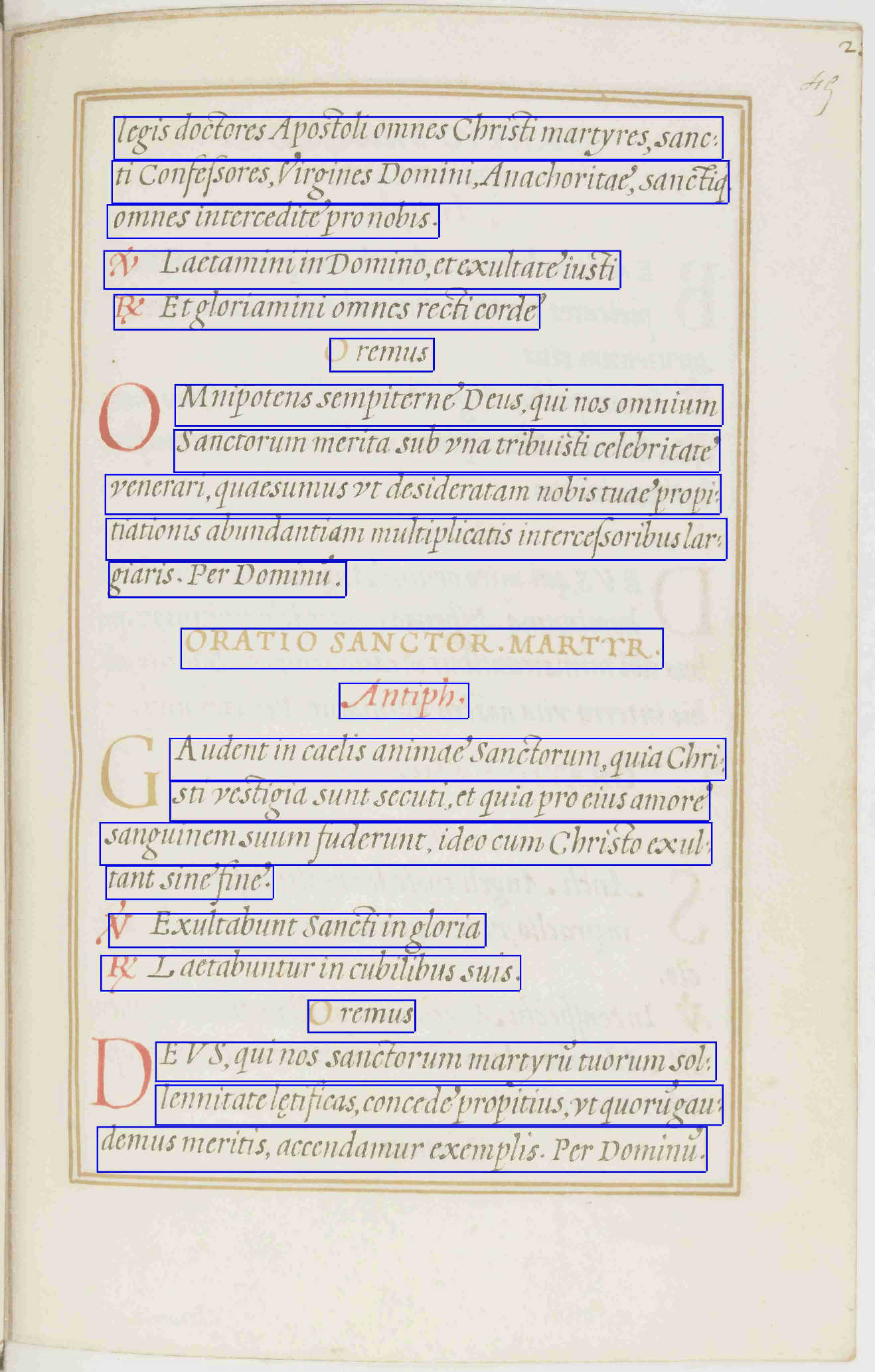}
    }
    \caption{Two pages from the Horae dataset with annotated text lines.}
    \label{fig:horae}
\end{figure}

Our main contributions are as follows. We propose Doc-UFCN, a U-shaped Fully Convolutional Network for text line detection. We show that this model outperforms a state-of-the-art method (dhSegment \cite{dhsegment}) while having less parameters and a reduced prediction time. In addition, we show that pre-training on documents images instead of natural scene images can increase the results even with few training data.

In this paper, we first describe (Section \ref{related_work}) the current advances in the field. Our model and it's implementation are detailed in Section \ref{model}. Then we present the data used (Section \ref{data}) and the experiments (Section \ref{comparison_SOTA}). Section \ref{ablation_study} is dedicated to an ablation study of our model.

%% file: parts/02-related_works.tex
In the recent years, the interest given to the analysis of historical documents has been boosted by the competitions on textline detection \cite{ICDAR2015}, baseline detection \cite{cBAD2017} \cite{cBAD2019}, layout analysis \cite{HDLAC2011} or writer identification \cite{WI2013}. There have been successful models and systems tackling historical document analysis problems such as text line segmentation or layout analysis.

Oliveria et al. \cite{dhsegment} recently proposed a convolutional neural network with an encoder pre-trained on ImageNet. This method has shown promising results on various tasks with few training data and the training time is significantly reduced due to the pre-trained encoder. This model differs from Doc-UFCN since its encoder follows the ResNet-50 \cite{resnet} achitecture and is pre-trained on natural scene images. Our encoder is smaller than dhSegment's, has way less parameters and is fully trained on document images. However, both models have similar decoders with the use of encoding feature maps during the decoding step.

Barakat et al. \cite{barakat2018} proposed a fully convolutional network for text line detection. Their architecture consists in successive convolutions and pooling layers during the encoding stage and upsampling layers in the decoding stage. Unlike us, they only use low level feature maps during the decoding step, upsampling them many times before combining them. This architecture has shown good results on Arabic handwritten pages but requires binarized input images. Mechi et al. \cite{mechi2019} presented an adaptive U-Net architecture for text line segmentation. Their encoder also consists in convolutions and pooling layers. During the decoding step, successive standard convolutions, transposed convolutions and sigmoid layers are applied. For our model, we also chose to use standard convolutions followed by transposed convolutions for the upsampling step. This allows to have the same resolution at both sides of the network.

Grüning et al. \cite{gruning2018} proposed a more complicated architecture composed of two stages to detect baselines in historical documents. First a hierarchical neural network (ARU-Net) is applied to detect the text lines. This ARU-Net is an extended version of the U-Net \cite{unet} architecture. A spatial attention network is incorporated to deal with various font sizes in pages. In addition, they added residual blocks to the U-Net architecture. This enables to train deeper neural networks while reaching higher results. Second, they apply successive steps to cluster superpixels to build baselines.

Yang et al. \cite{yang2018} designed a multimodal fully convolutional network for layout analysis. They take advantage of the text content as well as the visual appearance to extract the semantic structures of document images. This method has shown high Intersection-over-Union scores but requires more complex data annotations. Indeed, for each document image, a pixel-wise labeled image as well as textual contents are needed. Our model is based on the core architecture of this network. The use of dilated convolutions in the encoder allows to have a broader context information and more accurate results. Renton et al. \cite{renton2018} also demonstrated the advantages of using such convolutions instead of standard ones. Their fully convolutional network is composed of successive dilated convolutions that increase the receptive field. They are followed by one last standard convolution that outputs the labeled images.

Finally, Moysset et al. \cite{moysset2015} proposed a recurrent neural network to segment text paragraphs into text lines. This network differs from the systems presented above and from our because it has recurrent layers. It also differs since ground truth lines are not represented as bounding boxes but the paragraph itself is represented as a succession of line and interline labels.

%% file: parts/03-model.tex
Our goal is to analyse the impact of the pre-training step on the line segmentation task. To this aim, the proposed architectures are analyzed with and without pre-training. In this section, we detail two state-of-the-art architectures: dhSegment \cite{dhsegment} and Yang's \cite{yang2018}. We then present our model Doc-UFCN that is inspired by the core architecture proposed by Yang and give the implementation details.

\subsection{Comparison of architectures}
In this section, we detail the two architectures, dhSegment and Yang's one, and explain the choices we made to design our own model.

\subsubsection{dhSegment}
dhSegment is the state-of-the-art method for multiple layout analysis tasks on historical documents. It has shown various advantages like working with few training data and a reduced training time. In addition, the code to train and test the model is open-source\footnote{https://github.com/dhlab-epfl/dhSegment} and can be easily trained in the same conditions as our model to have a fair comparison.

\begin{figure*}
    \centering
    \begin{subfigure}[hbtp]{0.75\textwidth}
        \centerline{\includegraphics[width=1\textwidth]{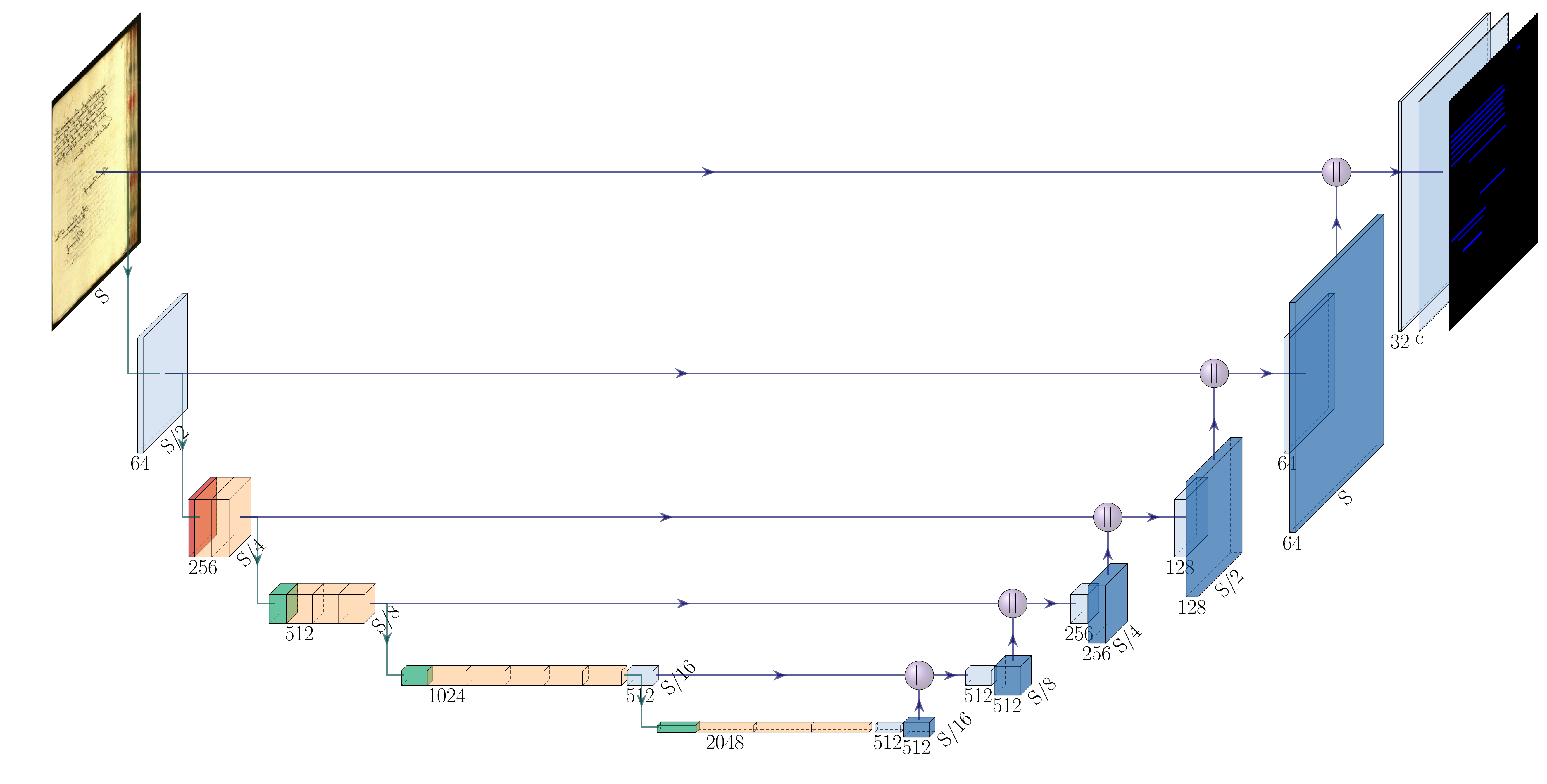}}
        \caption{Architecture of dhSegment. The encoding path corresponds to a modified version of the ResNet-50\cite{resnet} architecture.}
        \label{fig:dhsegment_archi}
    \end{subfigure}
    \begin{subfigure}[hbtp]{0.50\textwidth}
        \centerline{\includegraphics[width=1\textwidth]{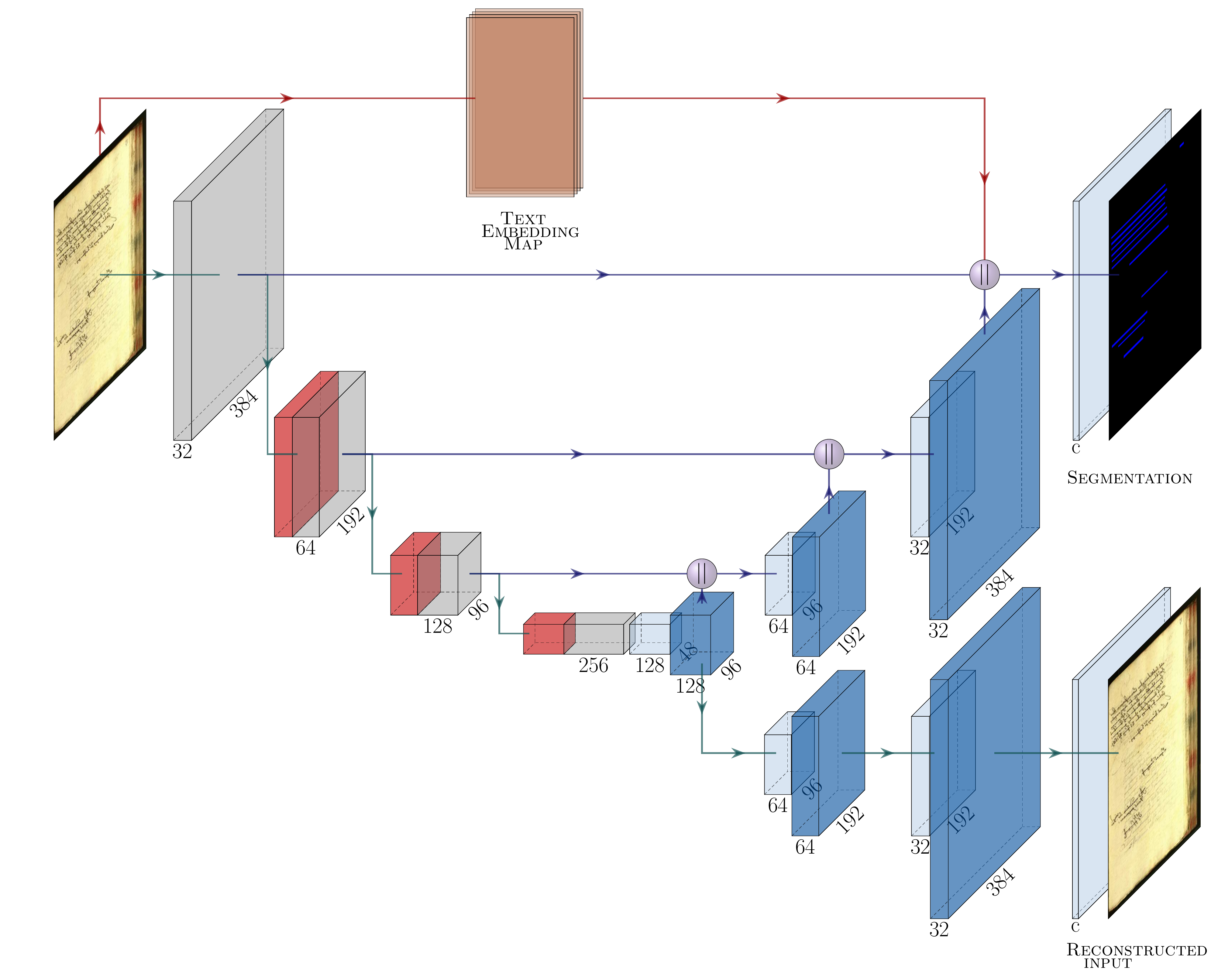}}
        \caption{Architecture of Yang's model.}
        \label{fig:yang_archi}
    \end{subfigure}
    \hspace{2cm}
    \begin{subfigure}[hbtp]{0.2\textwidth}
        \centerline{\includegraphics[width=1\textwidth]{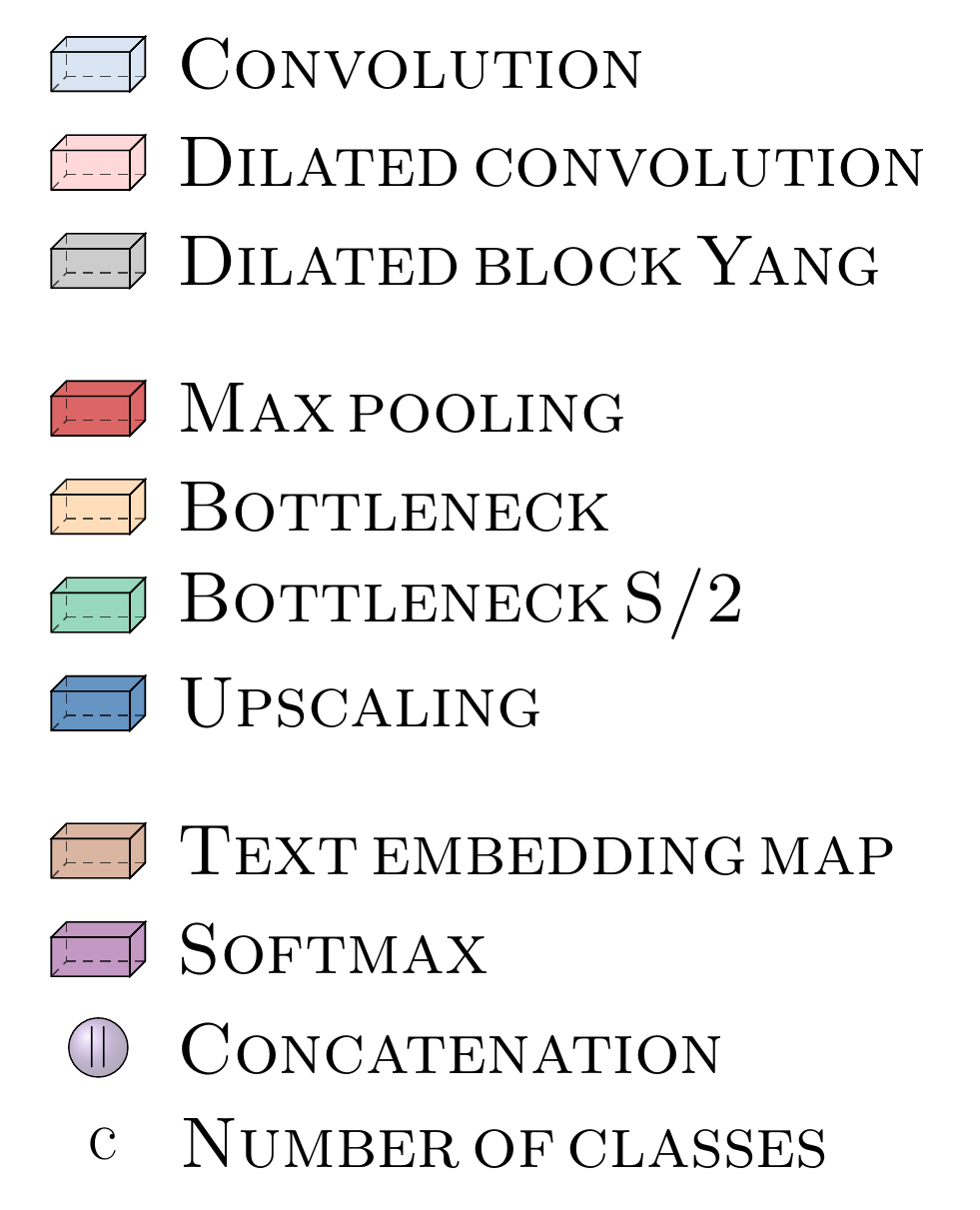}}
    \end{subfigure}
    \begin{subfigure}[hbtp]{0.65\textwidth}
        \centerline{\includegraphics[width=1\textwidth]{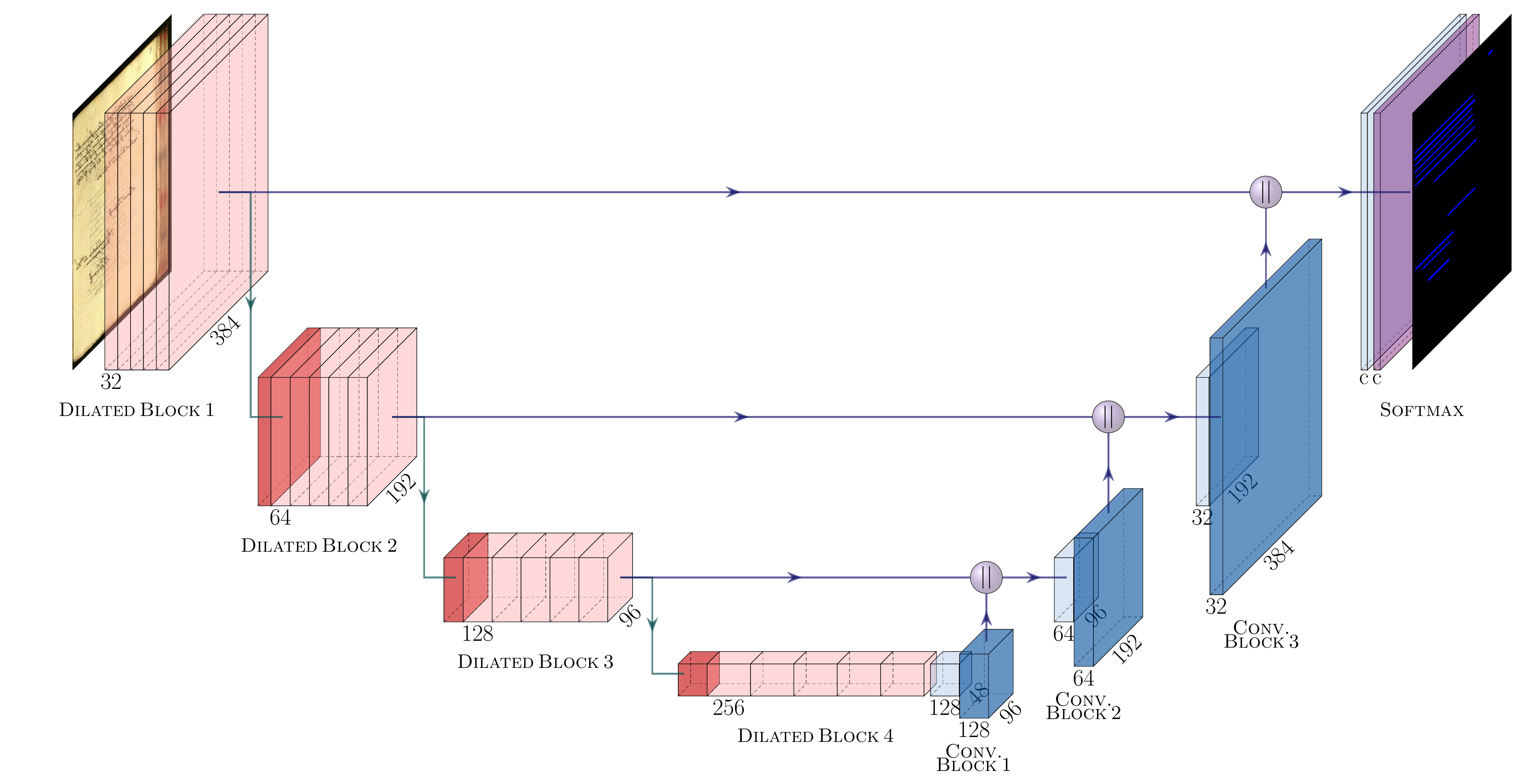}}
        \caption{Architecture of our model Doc-UFCN. The encoding path is represented in red and the decoding path in blue.}
        \label{fig:ufcn_archi}
    \end{subfigure}
    \caption{Architectures of the different models.}
\end{figure*}

dhSegment's architecture is presented Figure \ref{fig:dhsegment_archi}. This model is deeper than Yang's and can have up to \textit{2048} feature maps. The encoder is composed of convolution (light blue and orange on the Figure 2a) and pooling layers. This encoder is first pre-trained on natural scene images \cite{ImageNet} and both the encoder and decoder are then trained on document images. The decoder is quite similar to the one used by Yang and consists in successive blocks composed of one standard convolution and one upscaling layer.

\subsubsection{Yang et al.}
Yang's model is a multimodal fully convolutional network. It takes into account the visual and textual contents for the segmentation task. It has shown good performances on synthetic and real datasets of modern document images. The code to train the model is also open-source\footnote{http://personal.psu.edu/xuy111/projects/cvpr2017\_doc.html}.

Yang's model is presented on Figure \ref{fig:yang_archi}. It is made of 4 parts: an encoder (red blocks on the Figure \ref{fig:yang_archi}), a first decoder outputing a segmentation mask, a second decoder for the reconstruction task and a bridge (red arrows) used for the textual content. The Text Embedding Map and the bridge are used to encode the textual content of the images and then to add the text information to the visual one before the last convolution. To have a fair comparison with dhSegment, only the visual content is used. Therefore, the Text Embedding Map, the bridge and the second decoder for the reconstruction task are removed.

\subsection{Description of Doc-UFCN}

Recent systems can show long inference times which can have great financial and ecological impacts. Indeed, dhSegment takes up to 66 days to detect the lines of the whole Balsac corpus (almost 2 million pages) on a GeForce RTX 2070. To this aim, we want to show the impact of the pre-trained parts on the segmentation results while having a small network and a reduced prediction time. To design our model, we chose to use the core of Yang's network since it has a reduced number of parameters and contains no pre-trained parts. Therefore, our architecture is a Fully Convolutional Network (FCN) composed of an encoder (red blocks on the Figure \ref{fig:ufcn_archi}) followed by a decoder (blue blocks) and a final convolution layer. Dealing with a FCN without any dense layer has many advantages. First, it highly reduces the number of parameters since there is no dense connection. In addition, it allows the network to deal with variable input image size and to keep the spatial information as is.

To keep a light model, the second decoder used by Yang is not used in our architecture.

\subsubsection{Contracting path}
The contracting path (encoder) consists in 4 dilated blocks. The dilated blocks are slightly different from those presented by Yang et al. since they consist in 5 consecutive dilated convolutions. Using dilated convolutions instead of standard convolutions allows the receptive field to be larger and the network to have more context information. Each block is followed by a max-pooling layer except for the last one.

\subsubsection{Expanding path}
The goal of the expanding path (decoder) is to reconstruct the input image with a pixel-wise labeling at the original input image resolution. This deconvolution is usually done using transposed convolutions or upscaling. As suggested by Mechi et al. \cite{mechi2019}, we decided to replace the unpooling layers of Yang's model by transposed convolutions in order to keep the same resolution on both the input and output. Therefore, the decoding path is composed of 3 convolutional blocks, each consisting of a standard convolution followed by a transposed convolution. In addition, the features computed during the encoding step are concatenated with those of the decoding stage (purple arrows on the Figure \ref{fig:ufcn_archi}).

\subsubsection{Last convolution}
The last convolutional layer outputs full resolution feature maps. It returns \textit{c} feature maps with the same size as the input image, \textit{c} being the number of classes involved in the experiment. A softmax layer is then applied to transform these feature maps into probability maps.

\subsection{Implementation details}
We now present the implementation details of our model.

\subsubsection{Input image size}
Since our model is inspired by Yang et al. \cite{yang2018}, we decided to use the same input image size. We thus resized the input images and their corresponding label maps into smaller images of size \textit{384$\times$384 px}, adding padding to keep the original image ratio. This allows to reduce the training time without losing too much information. We also tested another input size to see the impact of this choice (see Section \ref{input_size}). 

\subsubsection{Dilated block}
As stated before, all the dilated blocks are composed of 5 consecutive dilated convolutions with dilation rates \textit{d = 1, 2, 4, 8} and \textit{16}. The blocks respectively have \textit{32, 64, 128} and \textit{256} filters. Each convolution has a \textit{3$\times$3} kernel, a stride of \textit{1} and an adapted padding to keep the same tensor shape throughout the block. All the convolutions of the blocks are followed by a Batch Normalization layer, a ReLU activation and a Dropout layer with a probability \textit{p\_dilated}.

\subsubsection{Convolutional block}
The convolutional blocks are used during the decoding step. The expanding path is composed of 3 convolutional blocks and each block is composed of a standard convolution followed by a transposed convolution. The blocks respectively have \textit{128, 64} and \textit{32} filters. Each standard convolution has a \textit{3$\times$3} kernel, a stride and a padding of \textit{1}. Each transposed convolution has a \textit{2$\times$2} kernel and a stride of \textit{2}. As for the dilated blocks, all the standard and transposed convolutions are followed by a Batch Normalization layer, a ReLU activation and a Dropout layer with a probability \textit{p\_conv}.

\subsubsection{Last convolution}
The last convolution layer is parametrized as follows: \textit{c} (number of classes) filters, \textit{3$\times$3} kernel, stride and padding of \textit{1}. It is followed by a softmax layer that computes the pixel's class conditional probabilities.

\subsubsection{Post-processing}
As a post-processing step, we apply the same operations as the one applied by dhSegment: pixels with a confidence score higher than a threshold \textit{t} are kept and connected components with less than \textit{min\_cc} pixels are removed.

%% file: parts/04-data.tex
The models have been tested on 4 datasets for the line segmentation task. Table \ref{tab:datasets} summarizes these datasets.

\begin{table}[htbp!]
    \begin{center}
        \begin{tabular}{ccccc}
            \hline
            \textsc{Dataset} & \textsc{Manuscripts} & \textsc{Pages} & \textsc{Lines} & \textsc{Mean size} \\ \hline
            Balsac & 74 & 913 & 46159 & [3882, 2418] \\
            Horae & 500 & 557 & 12431 & [4096, 5236] \\
            READ-BAD & 9 & 486 & 28066 & [4096, 2731] \\
            DIVA-HisDB & 3 & 120 & 11933 & [4992, 3328] \\
            \hline
        \end{tabular}
    \end{center}
    \caption{Details of the datasets: number of manuscripts from which the pages have been extracted, number of pages and annotated lines and mean size of the images.}
    \label{tab:datasets}
\end{table}

\begin{table*}[htbp!]
    \begin{center}
        \begin{tabular}{c|cc|cc|cc|cc}
            \hline
            \multirow{2}{*}{\textsc{Dataset}}&\multicolumn{2}{c}{\textsc{Mean IoU (\%)}}&\multicolumn{2}{|c}{\textsc{Precision (\%)}}&\multicolumn{2}{|c}{\textsc{Recall (\%)}}&\multicolumn{2}{|c}{\textsc{F1-score (\%)}} \\
            & dhSegment & Our & dhSegment & Our & dhSegment & Our & dhSegment & Our \\
            \hline
            Balsac & 73.78 & \textbf{83.79} & 92.07 & \textbf{94.80} & 78.76 & \textbf{87.86} & 84.81 & \textbf{91.11} \\ 
            Horae & \textbf{65.22} & 63.95 & 71.70 & \textbf{78.38} & \textbf{89.29} & 80.45 & 82.32 & \textbf{84.93} \\ 
            READ-Simple & \textbf{64.55} & 64.03 & \textbf{85.04} & 81.76 & 71.85 & \textbf{75.60} & \textbf{77.25} & 76.66 \\ 
            READ-Complex & 52.91 & \textbf{54.40} & 79.28 & \textbf{83.62} & 59.16 & \textbf{61.97} & 69.27 & \textbf{73.16} \\ 
            DIVA-HisDB & 74.24 & \textbf{75.71} &  \textbf{92.41} & 92.14 & 79.10 & \textbf{80.88} & 85.19 & \textbf{86.09} \\ 
            \hline
        \end{tabular}
    \end{center}
    \caption{Comparison of the results obtained by the two networks at pixel level.}
    \label{tab:results}
    \vspace{-0.2cm}
\end{table*}

\subsubsection{Balsac}

\begin{figure}[htbp]
    \centerline{
        \includegraphics[width=0.185\textwidth]{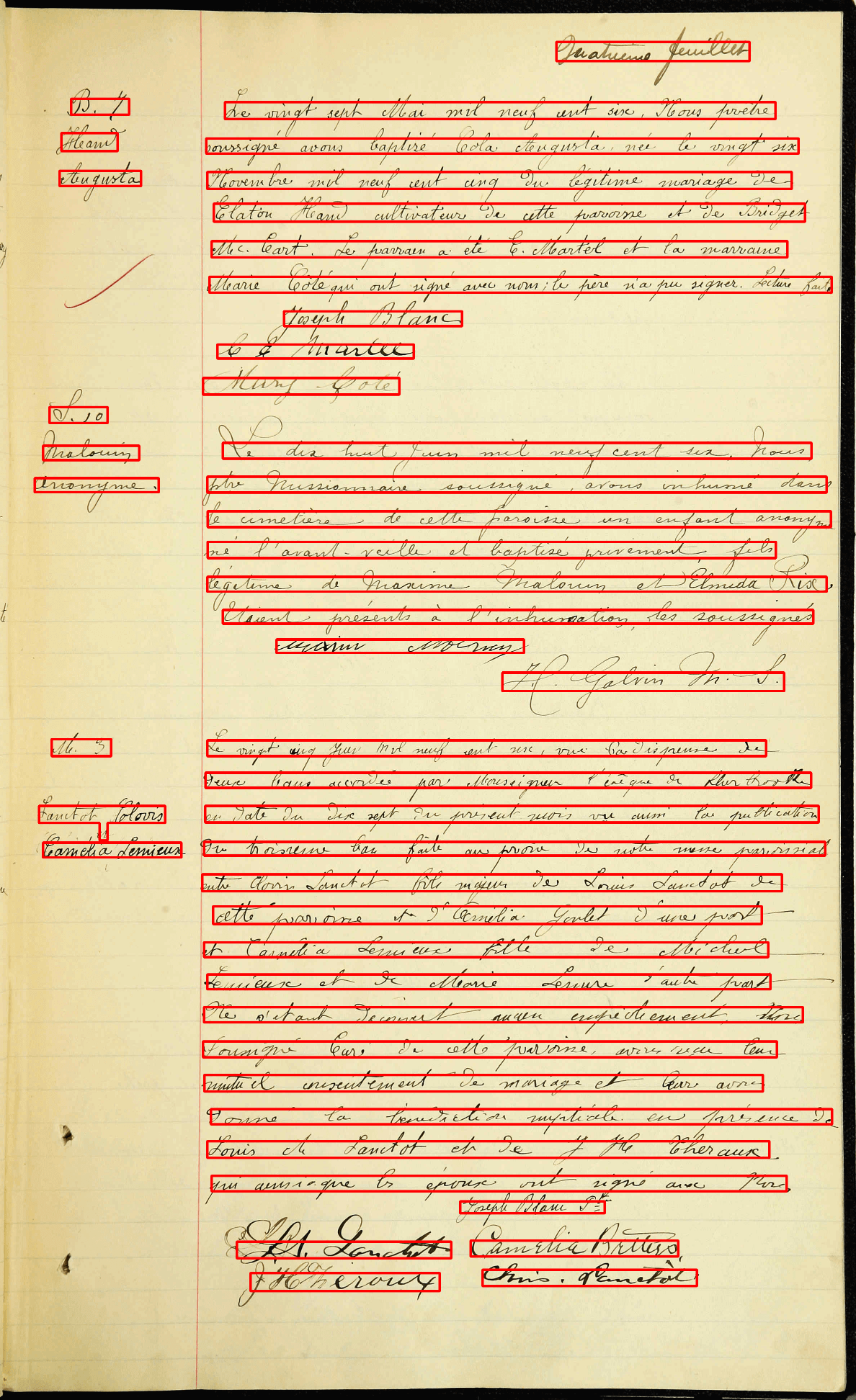}
        \hspace{0.1cm}
        \includegraphics[width=0.25\textwidth]{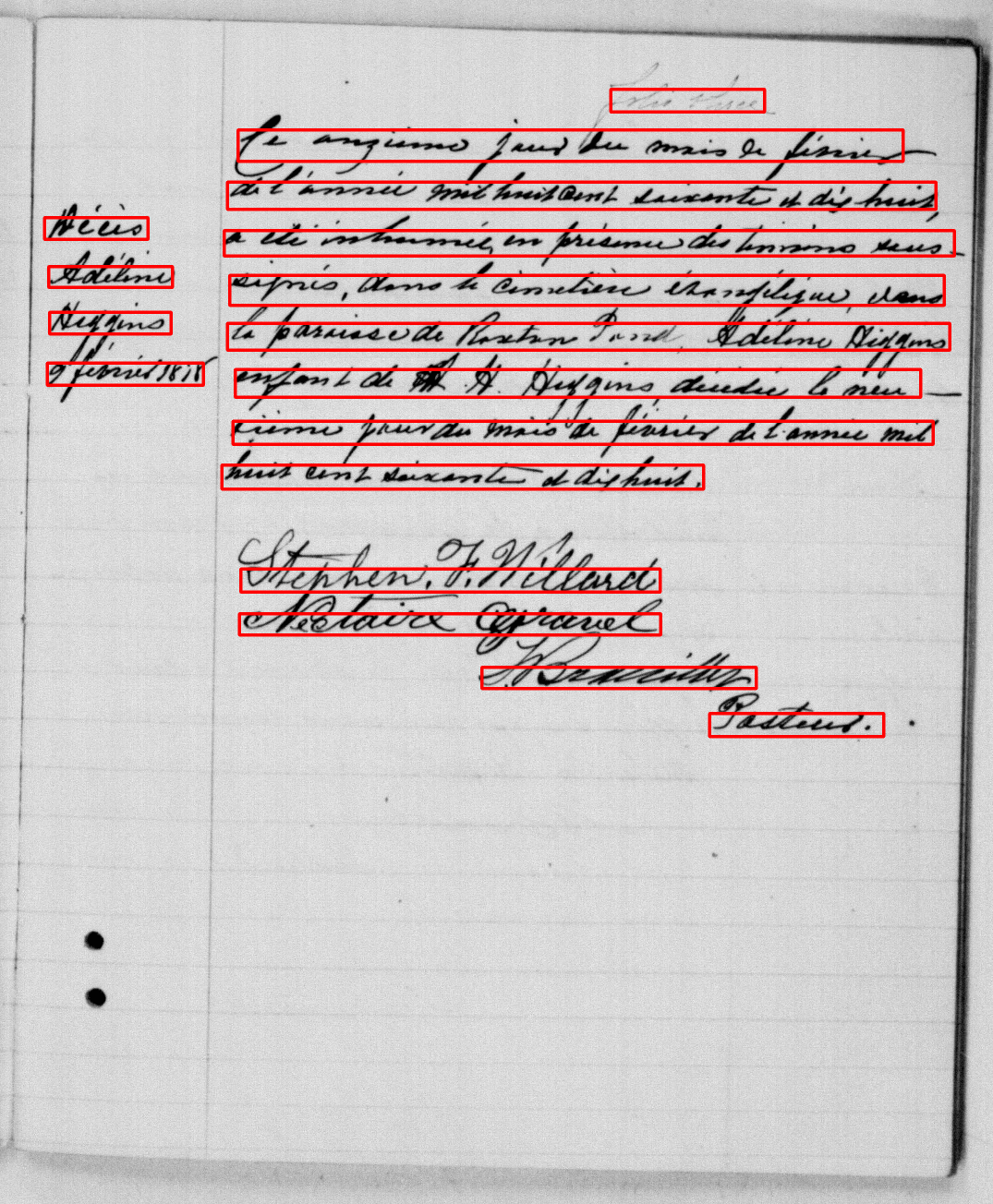}
    }
    \caption{Two pages from the Balsac dataset with annotated text lines. 3 full acts on the left and one full act on the right document image.} 
    \label{fig:balsac}
\end{figure}

The Balsac dataset consists in 913 images extracted from 74 registers selected among 44742 registers in total. The images represent pages of acts written in french and are annotated at line level. Two examples images are shown on Figure \ref{fig:balsac}.

\subsubsection{Horae}

This dataset consists in 557 annotated pages of books of hours. These pages have been selected among 500 manuscripts as they represent the variety of layouts and contents \cite{horae2019}. The pages have been annotated at different levels and with various classes such as simple initials, decorated initials or ornamentations. Figure \ref{fig:horae} shows two annotated pages for text line segmentation selected from two different manuscripts.

\subsubsection{READ-BAD}

This dataset \cite{gruning_cbad} is composed of 2036 annotated archival images of documents and has been used during the cBAD: ICDAR2017 \cite{cBAD2017} competition. The images have been extracted from 9 archives and the dataset is split into Simple and Complex subsets. Each image has its corresponding ground truth in PAGE xml format. For the line segmentation task, we used the bounding boxes of the \textit{TextLine} objects as labels.

\subsubsection{DIVA-HisDB}

This last dataset \cite{diva} contains 120 annotated pages extracted from 3 different manuscripts. Each manuscript has 30 training, 10 validation and 10 testing images.

%% file: parts/05-comparison_sota.tex
We applied Doc-UFCN to the 4 datasets. In addition, we also trained dhSegment architecture \cite{dhsegment} in the same conditions. In the following, we only compare our model to dhSegment since ours is too similar to Yang's to be compared with. This section details the trainings and shows the results obtained.

\subsection{Training}

Our model is implemented in PyTorch. We trained it with an initial learning rate of \textit{5e-3}, Adam optimizer and the cross entropy loss. The weights are initialized using Glorot initialization. In addition, we used mini-batches of size 4 to reduce the training time. We tested different dropout probabilities and decided to keep the model with \textit{p\_dilated = p\_conv = 0.4} since it yielded higher performances on average on the validation set. The model is trained over a maximum of 200 epochs and early stopping is used to stop training when the model converges. In the end, we keep the model with the lowest validation loss.

We also trained dhSegment on our data with the same splits for a maximum of 60 epochs since the model is pre-trained and converges faster than our. We used mini-batches of size 4 and trained on patches of shape \textit{400$\times$400 px}. The initial learning rate is \textit{5e-5} and we chose to use a ResNet50 \cite{resnet} as pre-trained encoder. Early stopping is also used and the best model obtained during training is selected.

Both models have the same post-processing step with the same hyper-parameters. After testing thresholds within a range from \textit{0.5} to \textit{0.9}, we kept \textit{t = 0.7} since it shows the best results on the validation set, allowing the expected pixels to be predicted as text lines and rejecting those belonging to the background. Lastly, the small connected components with less than \textit{min\_cc = 50} pixels are discarded. Several values have also been tested for this parameter, however, it didn't really impact the results obtained.

\subsection{Results}
\label{results}

We trained the two networks on the four datasets and now we report the scores obtained for both of them. Most of the existing methods are evaluated using the Intersection-over-Union (IoU) metric. The IoU measures the average similarity between the predicted and the ground truth pixels. Alberti et al. \cite{albertiDIVA} designed a tool to evaluate the performance of a model by calculating the IoU, precision, recall and F-measure. It allows to have more information concerning the model's performances at pixel level than just the IoU.

Therefore, to evaluate the models, we computed various pixel level metrics. We first report the Intersection-over-Union (IoU) as well as the Precision (P), Recall (R) and F1-score (F) in Table \ref{tab:results}. To be comparable, the images predicted by dhSegment are resized to \textit{384$\times$384 px} before computing the metrics. In addition, the values are only presented for the text line class (the background is not considered here).

The results obtained by our method are often better than those obtained by dhSegment. On the Balsac dataset, our model outperforms dhSegment by up to 6 percentage points for the F1-score metric. This is due to a better separation of close text lines that are often predicted as one single line by dhSegment. Our model helps separating these lines where dhSegment fails. It also helps to have smoother and more accurate contours.

So far, our model has shown better performances than dhSegment while having no pre-trained encoder. Another interesting point is that our model is way lighter than dhSegment. It has only 4.1M parameters to be learned whereas dhSegment has 32.8M parameters including 9.36M that have to be fully-trained. This leads to a reduced prediction time. Indeed, our model is up to 16 times faster than dhSegment model as shown on Table \ref{tab:prediction_time}. \\

\begin{table}[htbp!]
    \begin{center}
        \begin{tabular}{c|cc|c}
            \hline
            \multirow{2}{*}{\textsc{Dataset}}&\multicolumn{2}{c|}{\textsc{Mean prediction time$^1$}}&\multirow{2}{*}{\textsc{Ratio}} \\
            & dhSegment & Our & \\
            \hline
            Balsac & 2.95 & 0.41 & 7.20 \\
            Horae & 7.87 & 0.97 & 8.11 \\
            READ-Simple & 3.73 & 0.45 & 8.29 \\
            READ-Complex & 4.70 & 0.59 & 7.97 \\
            DIVA-HisDB & 12.90 & 0.80 & 16.13 \\
            \hline
            \multicolumn{4}{l}{\footnotesize{$^1$ Predictions made on a GPU GeForce RTX 2070 8G.}} \\[-0.1ex]
        \end{tabular}
    \end{center}
    \caption{Prediction times (s / image) reported for the two networks for the experiments presented in Section \ref{results}. The ratio column contains the improvement ratios (dhSegment / our times).}
    \label{tab:prediction_time}
\end{table}

\begin{table*}[htbp]
    \begin{center}
        \begin{tabular}{c|cc|cc|cc|cc}
            \hline
            \multirow{2}{*}{\textsc{Dataset}}&\multicolumn{2}{c}{\textsc{Mean IoU (\%)}}&\multicolumn{2}{|c}{\textsc{Precision (\%)}}&\multicolumn{2}{|c}{\textsc{Recall (\%)}}&\multicolumn{2}{|c}{\textsc{F1-score (\%)}} \\
            & dhSegment & Our & dhSegment & Our & dhSegment & Our & dhSegment & Our \\
            \hline
            Balsac & 74.02 & \textbf{84.87} & 91.89 & \textbf{94.25} & 79.09 & \textbf{89.49} & 84.95 & \textbf{91.75} \\
            Horae & 60.69 & \textbf{68.81} & \textbf{80.94} & 80.31 & 73.65 & \textbf{84.80} & 81.99 & \textbf{88.62} \\
            READ-Simple & 65.07 & \textbf{68.14} & \textbf{88.34} & 83.19 & 71.56 & \textbf{78.05} & \textbf{80.72} & 79.45 \\
            READ-Complex & 53.34 & \textbf{60.28} & \textbf{85.51} & 81.03 & 57.80 & \textbf{68.17} & 68.47 & \textbf{78.30} \\
            DIVA-HisDB & 73.00 & \textbf{74.72} & \textbf{91.56} & 89.43 & 78.28 & \textbf{82.20} & 84.32 & \textbf{85.44} \\
            \hline
            Balsac Fine-tuning & 74.52 & \textbf{85.73} & 91.48 & \textbf{92.90} & 80.03 & \textbf{91.70} & 85.29 & \textbf{92.24} \\
            Horae Fine-tuning & 62.79 & \textbf{68.00} & \textbf{86.91} & 79.51 & 71.12 & \textbf{84.51} & 79.91 & \textbf{87.97} \\
            READ-Simple Fine-tuning & 64.39 & \textbf{68.14} & \textbf{86.22} & 83.19 & 71.29 & \textbf{78.05} & 77.39 & \textbf{79.45} \\
            READ-Complex Fine-tuning & 52.96 & \textbf{60.28} & \textbf{85.63} & 81.03 & 57.43 & \textbf{68.17} & 68.95 & \textbf{78.30} \\
            DIVA-HisDB Fine-tuning & 74.24 & \textbf{74.72} & \textbf{92.83} & 89.43 & 78.79 & \textbf{82.20} & 85.18 & \textbf{85.44} \\
            \hline
        \end{tabular}
    \end{center}
    \caption{Comparison of the results obtained by the two networks at pixel level. The two models have been trained on the \textit{Multiple document dataset}. In the second part of the table, the models have been fine-tuned on the corresponding dataset.}
    \label{tab:results_pre-training}
\end{table*}

\subsection{Pre-training}

We have shown that pre-training on natural scene images is not required to have good results on document images. It is sometimes even worse than having a different model without any pre-trained part. We now want to see if pre-training on document images instead of natural scene images can have a positive impact on the performances. Therefore, in addition to the previous experiments, we trained dhSegment and our model on a mixture of all the datasets presented before. This dataset is denoted in the following as the \textit{Multiple document dataset}. The splitting obtained by mixing these images is shown in Table \ref{tab:split}.

\begin{table}[htbp]
    \begin{center}
        \begin{tabular}{lcc}
            \hline
            & \sc{Pages} & \sc{Text lines} \\ \hline
            \sc{Train} 		& \textbf{1688}  & \textbf{77454}  \\
            \hspace{0.4cm} Balsac & \hspace{0.4cm} 730 & \hspace{0.4cm} 37191 \\
            \hspace{0.4cm} Horae & \hspace{0.4cm} 510 & \hspace{0.4cm} 11341 \\
            \hspace{0.4cm} READ-Simple & \hspace{0.4cm} 172 & \hspace{0.4cm} 5117 \\
            \hspace{0.4cm} READ-Complex & \hspace{0.4cm} 216 & \hspace{0.4cm} 17768 \\
            \hspace{0.4cm} DIVA-HisDB & \hspace{0.4cm} 60 & \hspace{0.4cm} 6037 \\
            \sc{Validation}	& \textbf{188}  & \textbf{10562}  \\
            \hspace{0.4cm} Balsac & \hspace{0.4cm} 92 & \hspace{0.4cm} 4612 \\
            \hspace{0.4cm} Horae & \hspace{0.4cm} 17 & \hspace{0.4cm} 251 \\
            \hspace{0.4cm} READ-Simple & \hspace{0.4cm} 22 & \hspace{0.4cm} 540 \\
            \hspace{0.4cm} READ-Complex & \hspace{0.4cm} 27 & \hspace{0.4cm} 2160 \\
            \hspace{0.4cm} DIVA-HisDB & \hspace{0.4cm} 30 & \hspace{0.4cm} 2999 \\
            \sc{Test} 		& \textbf{200} & \textbf{10573} \\
            \hspace{0.4cm} Balsac & \hspace{0.4cm} 91 & \hspace{0.4cm} 4356 \\
            \hspace{0.4cm} Horae & \hspace{0.4cm} 30 & \hspace{0.4cm} 839 \\
            \hspace{0.4cm} READ-Simple & \hspace{0.4cm} 22 & \hspace{0.4cm} 723 \\
            \hspace{0.4cm} READ-Complex & \hspace{0.4cm} 27 & \hspace{0.4cm} 1758 \\
            \hspace{0.4cm} DIVA-HisDB & \hspace{0.4cm} 30 & \hspace{0.4cm} 2897 \\ \hline
            \sc{Total} 		& \textbf{2076} & \textbf{98589} \\ \hline
        \end{tabular}
    \end{center}
    \caption{Splitting of the \textit{Multiple document dataset}.}
    \label{tab:split}
\end{table}

\begin{figure}[htbp]
    \centerline{
        \includegraphics[width=0.16\textwidth]{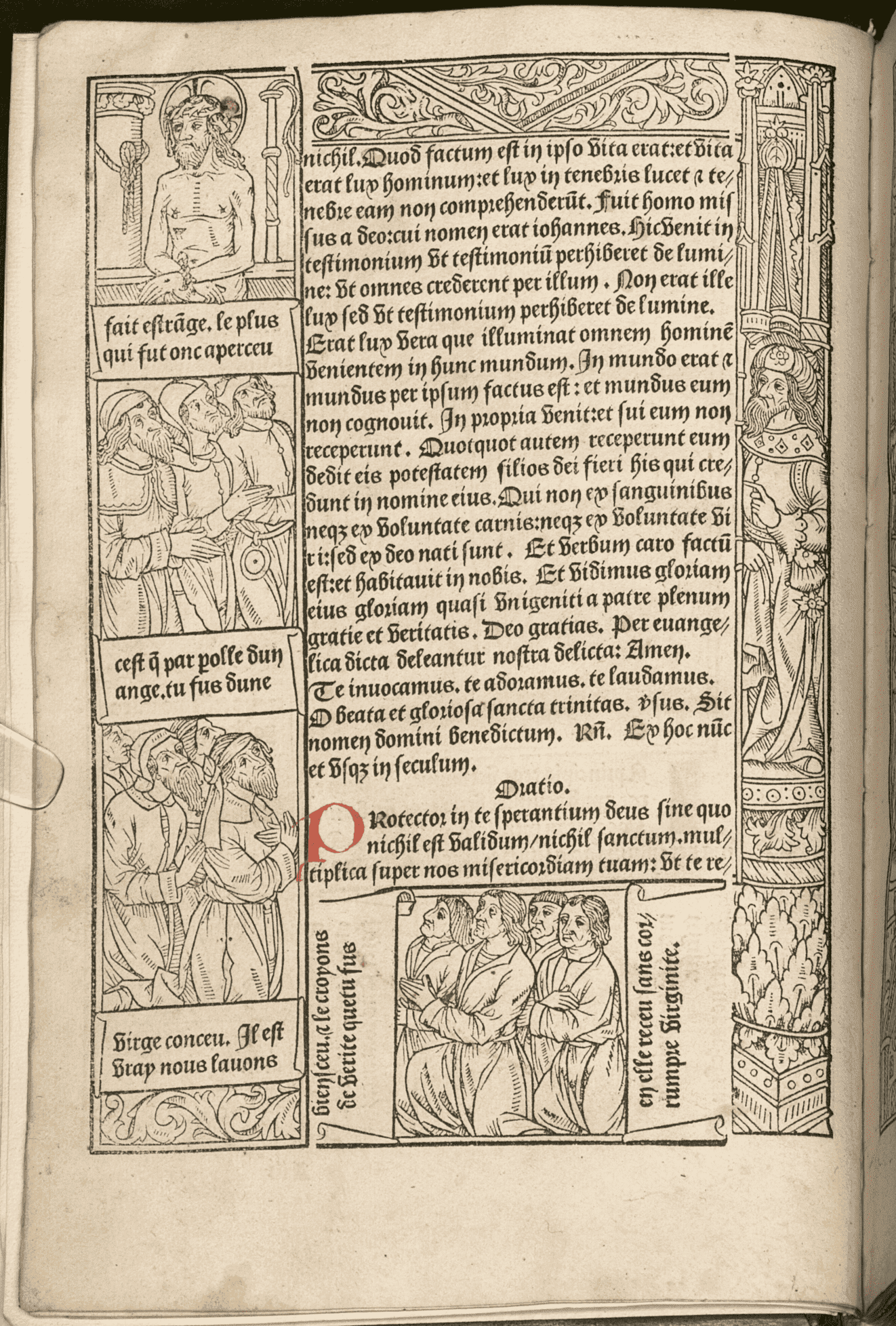}
        \includegraphics[width=0.16\textwidth]{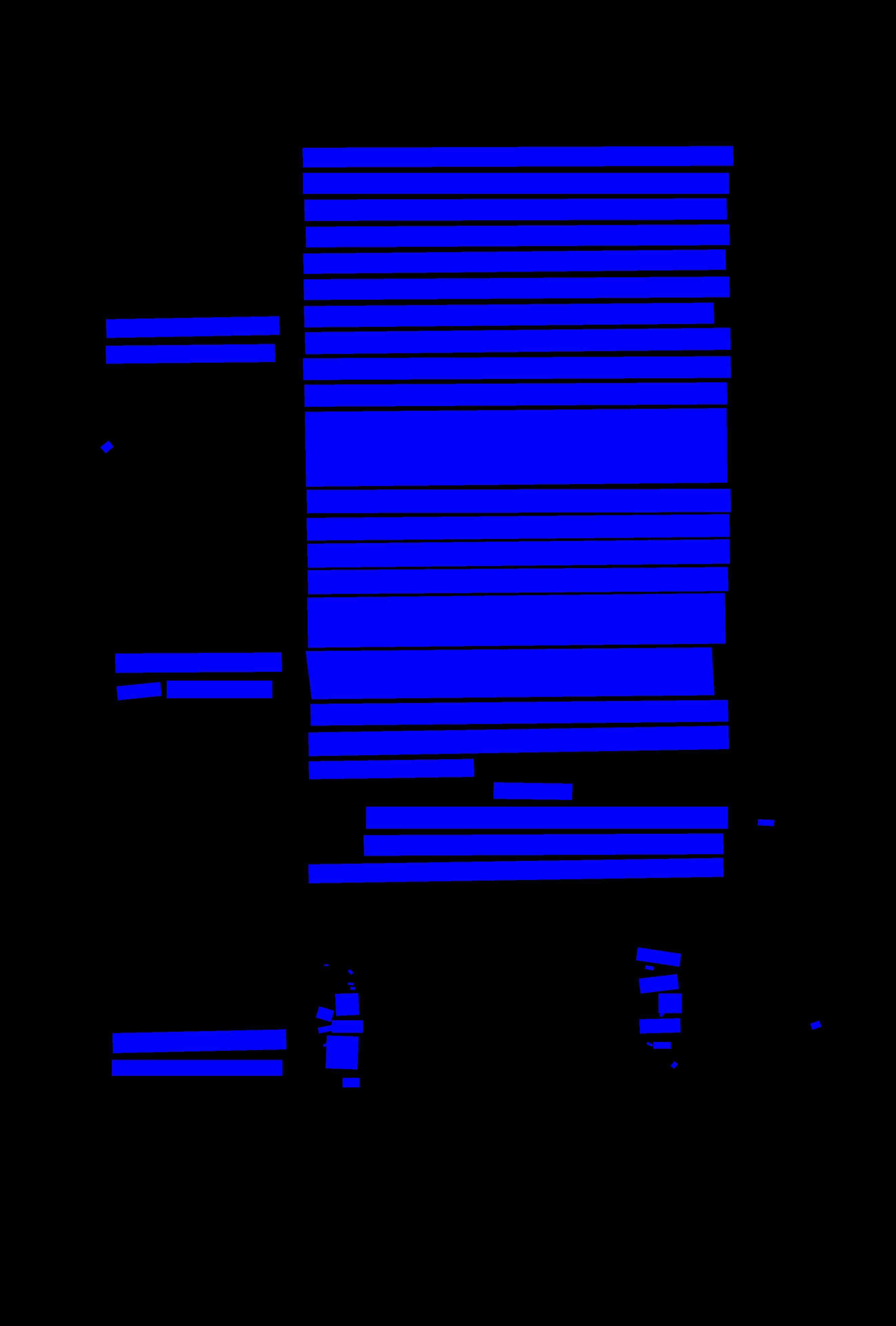}
        \includegraphics[width=0.16\textwidth]{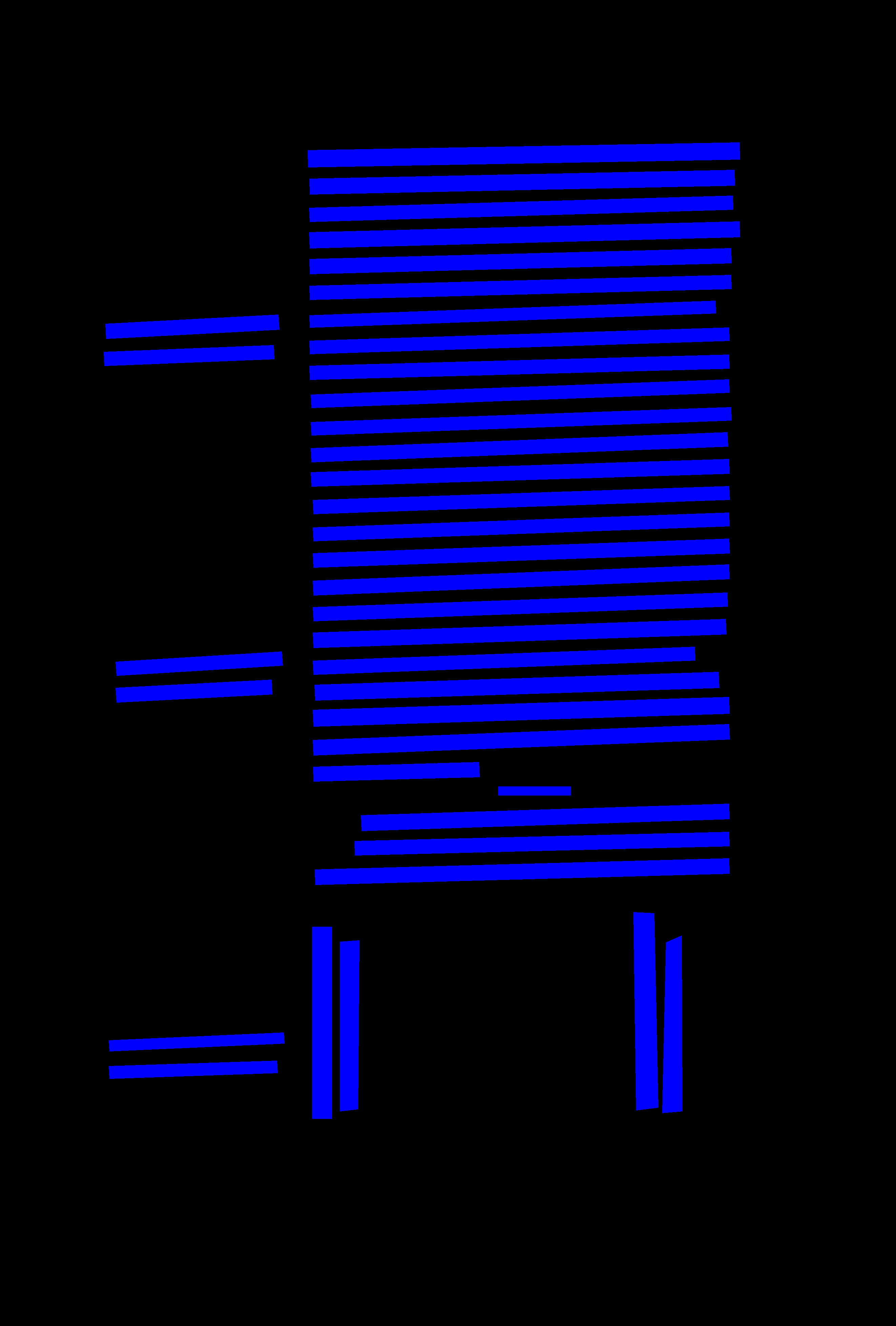}
    }
    \caption{Page from Horae dataset with the results of a line segmentation made by dhSegment (middle) and our model (right). dhSegment merges some lines and fails in detecting vertical text lines where our model correctly detect them.}
    \label{fig:visual_results}
\end{figure}

These generic models have then been tested on each dataset. The results are reported in Table \ref{tab:results_pre-training}. We also fine-tuned the models on each single dataset. To do so, we continued the training of our model for 80 epochs and dhSegment for 40 epochs.

Without any fine-tuning, our architecture is almost always better than dhSegment's. One can see that our architecture lacks in precision indicating that our model sometimes predicts text line pixels that belong to the background. However, recall is higher than dhSegment's which indicates that more of the expected text line pixels are found. This is more interesting for us since it means that we don't miss any characters. Figure \ref{fig:visual_results} shows the results of the two models for an image from Horae dataset.

Fine-tuning on each single dataset is not required to get good results with any of the models. With our architecture, only the model trained on Balsac took advantage of this fine-tuning. For the three other datasets, fine-tuning didn't improve the results since the best model obtained remains the one before re-training.

These results show that our model is better than dhSegment whatever the dataset, with and without fine-tuning. Adding this pre-training step to our model has improved the results, mostly on the READ datasets. This impact is less important on Balsac, mainly because this dataset represents 43 \% of the \textit{Multiple document dataset}. DIVA-HisDB is also less impacted by the pre-training. This is due to the small quantity of training data it has and the high complexity of the pages.

%% file: parts/06-ablation_study.tex
We did additional experiments with our model in order to see the impact of some components as well as external factors such as the size of the training set or the input image size. Table \ref{tab:ablation} summarizes the results obtained and the next sections describe the tested configurations.

\begin{table}[htbp!]
    \begin{center}
        \begin{tabular}{c|c|c|c|c|c}
            \hline
           \textsc{Dataset}&\sc{Version}&\textsc{IoU (\%)}&\textsc{P (\%)}&\textsc{R (\%)}&\textsc{F (\%)} \\
            \hline
            \multirow{4}{*}{Balsac} & $\emptyset$ & 78.91 & 95.27 & 82.11 & 88.13 \\
            & BN & 80.58 & \textbf{95.69} & 83.58 & 89.16 \\
            & BN + Drop1 & 83.40 & 94.16 & 87.95 & 90.87 \\
            & BN + Drop2 & \textbf{84.33} & 92.49 & \textbf{90.49} & \textbf{91.42} \\
            \hline
            \multirow{4}{*}{Horae} & $\emptyset$ & 56.02 & 81.94 & 67.37 & 77.85 \\
            & BN & \textbf{64.20} & 72.75 & \textbf{87.17} & 80.54 \\
            & BN + Drop1 & 63.98 & \textbf{84.60} & 74.76 & 83.17 \\
            & BN + Drop2 & 63.95 & 78.38 & 80.45 & \textbf{84.93} \\ \hline
            \multirow{4}{*}{READ-Simple} & $\emptyset$ & 58.85 & 78.89 & 69.15 & 72.59 \\
            & BN & 58.35 & \textbf{85.63} & 65.81 & 72.39 \\
            & BN + Drop1 & \textbf{66.34} & 81.64 & \textbf{79.14} & \textbf{78.08} \\
            & BN + Drop2 & 64.03 & 81.76 & 75.60 & 76.66 \\ \hline
            \multirow{4}{*}{READ-Complex} & $\emptyset$ & 39.38 & \textbf{88.04} & 42.37 & 56.41 \\
            & BN & 49.90 & 81.71 & 58.25 & 68.98 \\
            & BN + Drop1 & 51.87 & 86.73 & 56.58 & 68.74 \\
            & BN + Drop2 & \textbf{54.40} & 83.62 & \textbf{61.97} & \textbf{73.16} \\ \hline
            \multirow{4}{*}{DIVA-HisDB} & $\emptyset$ & 40.61 & 86.02 & 44.53 & 55.92 \\
            & BN & 73.57 & 91.85 & 78.80 & 84.69 \\
            & BN + Drop1 & 74.24 & 91.35 & 79.81 & 85.09 \\
            & BN + Drop2 & \textbf{75.71} & \textbf{92.14} & \textbf{80.88} & \textbf{86.09} \\
            \hline
        \end{tabular}
    \end{center}
    \caption{Comparison of the results obtained by different versions of our network (BN = Batch Normalization).}
    \label{tab:ablation}
\end{table}

\subsection{Batch Normalization}

As stated in \cite{batchnorm}, Batch Normalization has a great impact on the convergence speed during training but can also impact the results. Indeed, our model converged more than twice faster with Batch Normalization. In addition, as shown in Table \ref{tab:ablation}, Batch Normalization has a real impact on the F1-score in particular for Horae, READ-Complex and DIVA-HisDB. In addition to the quantitative results, we remarked that the visual results with Batch Normalization are also improved. It helps separating close regions but also helps joining regions that would be separated otherwise. In addition, the contours of the predicted regions are often more accurate and smoother.

\subsection{Dropout}

We tested two configurations with dropout layers. The first one (\textit{Drop1}) consists in applying a dropout with \textit{p\_dilated = p\_conv = 0.4} only after the dilated blocks. The second one (\textit{Drop2}) consists in applying the same dropout after every convolution of the model and not only after the last one of the dilated blocks. The application of dropout layers has most of the time a good impact on the performances. Even if the first configuration gives better results on the Horae and READ-Simple datasets, the impact is greater when implemented using the second configuration.

\subsection{Dilation}

For implementing the model, we chose to use a modified version of the dilated block proposed by Yang et al. \cite{yang2018} to have more context information to predict the text lines. To justify our choice of dilation rates, we tested 4 different configurations on the Balsac dataset. We tested blocks with only one convolution and a dilation rate of 1 ($[$\textit{1}$]$) and blocks with a dilation rate of 16 ($[$\textit{16}$]$). We also tested blocks with 5 convolutions with different rates ($[$\textit{1, 1, 1, 1, 1}$]$ and $[$\textit{1, 2, 4, 8, 16}$]$). The results obtained are presented in Table \ref{tab:dilation}.

\begin{table}[htbp!]
    \begin{center}
        \begin{tabular}{c|c|c|c|c}
            \hline
           \sc{Dilation}&\textsc{IoU (\%)}&\textsc{P (\%)}&\textsc{R (\%)}&\textsc{F (\%)} \\
            \hline
            $[$\textit{1}$]$ & 75.94 & \textbf{95.02} & 79.07 & 86.20 \\
            $[$\textit{1, 1, 1, 1, 1}$]$ & 79.93 & 92.02 & 85.77 & 88.57 \\
            $[$\textit{16}$]$ & 77.45 & 91.68 & 83.22 & 87.13 \\
            $[$\textit{1, 2, 4, 8, 16}$]$ & \textbf{83.79} & 94.80 & \textbf{87.86} & \textbf{91.11} \\
            \hline
        \end{tabular}
    \end{center}
    \caption{Impact of the dilation rates.}
    \label{tab:dilation}
\end{table}

\begin{figure}[htbp]
    \centerline{
        \includegraphics[width=0.50\textwidth]{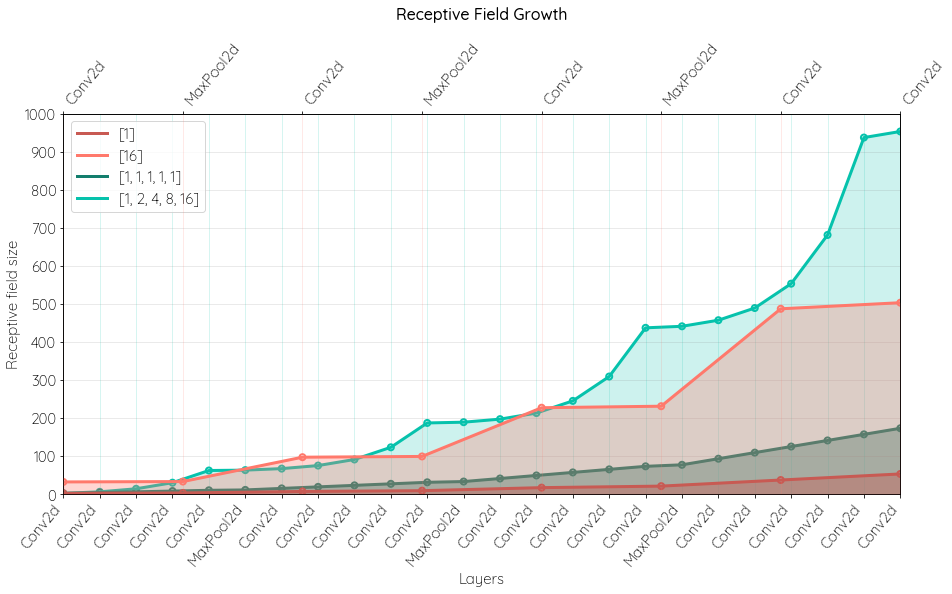}
    }
    \caption{Receptive field growth through the network.} 
    \label{fig:receptive_field}
\end{figure}

The results with the last configuration are better than any of the others since the receptive field is way larger and the model has more context to predict the text lines. Figure \ref{fig:receptive_field} shows the receptive field growth through the network. The receptive field with the dilation rate ($[$\textit{16}$]$) corresponds to the one of Yang's model since the dilated convolutions are not successive. Having dilated convolutions instead of standard ones really impacts the receptive field size (1000 pixels instead of 200) which results in using more context to predict the text lines and provides higher performances.

\subsection{Training set size}

In addition to the ablation study, we tried to analyze the impact of the training set size on the performances. Therefore, we trained our model on 4 subsets of Balsac training set and report the results on Table \ref{tab:training_size}.

\begin{table}[htbp!]
    \begin{center}
        \begin{tabular}{c|c|c|c|c}
            \hline
           \sc{Number of images}&\textsc{IoU (\%)}&\textsc{P (\%)}&\textsc{R (\%)}&\textsc{F (\%)} \\
            \hline
            90 (12\%) & 77.42 & 92.40 & 82.52 & 87.00 \\
            182 (25\%) & 78.64 & 95.17 & 81.85 & 87.91 \\
            365 (50\%) & 80.58 & \textbf{95.69} & 83.58 & 89.16 \\
            731 (100\%) & \textbf{81.95} & 94.53 & \textbf{85.92} & \textbf{89.89} \\
            \hline
        \end{tabular}
    \end{center}
    \caption{Impact of the training set size.}
    \label{tab:training_size}
\end{table}

The more the training data, the higher the IoU. However, this progression doesn't have the same effect on the precision metric. The model trained with 365 images has even a higher precision value than the one trained with 731 images. Moreover, we see that training over only 90 images (12 \% of the training set) gives quite good results which are even better than those obtained by dhSegment when trained on the whole dataset. 

\subsection{Input image size}
\label{input_size}

As we wanted to follow the model proposed in \cite{yang2018}, we decided to train our models on images resized to \textit{384$\times$384 px}. However we want to see the impact of this choice on our results. Therefore, we trained a model on Balsac and one on DIVA-HisDB on images resized to \textit{768$\times$768 px}. Table \ref{tab:image_size} shows that training on larger images improves a bit the results. However this impact is bigger when the training set contains a lot of images. Balsac dataset contains 731 training images and is more impacted than the DIVA-HisDB dataset that contains only 60 training images.

\begin{table}[htbp!]
    \begin{center}
        \begin{tabular}{c|c|c|c|c|c}
            \hline
            \sc{Dataset}&\sc{Size}&\textsc{IoU (\%)}&\textsc{P (\%)}&\textsc{R (\%)}&\textsc{F (\%)} \\
            \hline
            \multirow{2}{*}{Balsac} & 384 & 83.79 & \textbf{94.80} & 87.86 & 91.11 \\
            & 768 & \textbf{86.50} & 94.57 & \textbf{91.06} & \textbf{92.69} \\
            \hline
            \multirow{2}{*}{DIVA-HisDB} & 384 & 75.71 & 92.14 & 80.88 & 86.09 \\
            & 768 & \textbf{76.55} & \textbf{93.33} & \textbf{80.98} & \textbf{86.67} \\
            \hline
        \end{tabular}
    \end{center}
    \caption{Impact of the input image size.}
    \label{tab:image_size}
\end{table}

%% file: parts/07-conclusion.tex
In this paper, we introduced a new model Doc-UFCN to detect the text lines from historical document images. This model takes advantage of a lot of context information due to the dilated convolutions whereas most of the existing methods only use standard ones. Moreover, it doesn't use any pre-trained weights learned on natural scene images but has shown better performances than state-of-the-art model.

We showed that there is no need to use heavy pre-trained encoders like ResNet. Using a different architecture like ours can give better results while being lighter than dhSegment, working with less training images and having a reduced prediction time. We also showed that pre-training a simple architecture on few document images improves the line detection. We don't need a huge amount of data to have a good pre-trained network.

Our future works will consist in evaluating our model on other tasks like the act segmentation of Balsac pages and the layout analysis of Horae images.